%% file: root.tex
\documentclass[letterpaper, 10 pt, conference]{ieeeconf}  

\input{includes}

\IEEEoverridecommandlockouts                              

\usepackage{tabularx} 

\title{\Large \bf
PRISM-DP: Spatial Pose-based Observations for Diffusion-Policies via Segmentation, Mesh Generation, and Pose Tracking
}

\author{Xiatao Sun, Yinxing Chen, Daniel Rakita
\thanks{All authors are with the Department of Computer Science, Yale University,
        New Haven, CT 06520, USA
        {\tt\small \{xiatao.sun, j.y.chen, daniel.rakita\}@yale.edu}}
}     

\begin{document}

\maketitle
\thispagestyle{empty}
\pagestyle{empty}


\input{0-abstract}
\input{1-introduction}

\input{2-related_works}

\input{3-technical_overview}

\input{5-evaluation}

\input{6-discussion}





\bibliographystyle{plainnat}

\bibliography{refs}


\end{document}

%% file: includes.tex
\usepackage{times}
\usepackage[numbers]{natbib}
\usepackage{multicol}
\usepackage[bookmarks=true]{hyperref}

\usepackage{lipsum}
\usepackage{comment}
\usepackage{balance}
\usepackage{graphicx}
\usepackage[table]{xcolor}
\usepackage{caption}
\usepackage{listings}
\usepackage{xcolor}

\definecolor{codegreen}{rgb}{0,0.6,0}
\definecolor{codegray}{rgb}{0.5,0.5,0.5}
\definecolor{codepurple}{rgb}{0.58,0,0.82}
\definecolor{backcolour}{rgb}{0.95,0.95,0.92}

\lstdefinestyle{mystyle}{
    backgroundcolor=\color{backcolour},
    commentstyle=\color{codegreen},
    keywordstyle=\color{magenta},
    numberstyle=\tiny\color{codegray},
    stringstyle=\color{codepurple},
    basicstyle=\ttfamily,
    breaklines=true,
    captionpos=b,
    keepspaces=true,
    numbers=left,
    numbersep=5pt,
    showstringspaces=false,
    tabsize=2
}
\usepackage{minted}
\usepackage{xcolor}
\definecolor{bg}{HTML}{f2f2ea}
\usepackage{float} 

\usepackage{seqsplit}

\usepackage{amssymb}
\usepackage{amsmath}
\usepackage{booktabs}
 
\usepackage{graphicx} 
\usepackage{subcaption}

%% file: 0-abstract.tex
\begin{abstract}
Diffusion policies generate robot motions by learning to denoise action-space trajectories conditioned on observations. These observations are commonly streams of RGB images, whose high dimensionality includes substantial task-irrelevant information, requiring large models to extract relevant patterns. In contrast, using structured observations like the spatial poses of key objects enables training more compact policies with fewer parameters. However, obtaining accurate object poses in open-set, real-world environments remains challenging, as 6D pose estimation and tracking methods often depend on markers placed on objects beforehand or pre-scanned object meshes that require manual reconstruction. We propose PRISM-DP, an approach that leverages segmentation, mesh generation, and pose tracking models to enable compact diffusion policy learning directly from the spatial poses of task-relevant objects. Crucially, by using a mesh generation model, PRISM-DP eliminates the need for manual mesh creation, improving scalability in open-set environments. Experiments in simulation and the real world show that PRISM-DP outperforms high-dimensional image-based policies and achieves performance comparable to policies trained with ground-truth state information.
\end{abstract}

%% file: 1-introduction.tex
\section{Introduction}
\label{sec:introduction}

Diffusion-based visuomotor policies generate robot motions by learning to denoise action-space trajectories conditioned on observations \cite{chi2023diffusion}.  These observations are commonly sequences of raw RGB images, appealing due to their generality and ease of deployment, requiring just a camera.  However, raw image sequences contain substantial amounts of irrelevant information, making it challenging for models to extract meaningful patterns without extensive computational resources, often necessitating hundreds of millions of parameters to fit demonstration data effectively.

In contrast, compact diffusion policies can be trained when using more structured observations, such as the spatial poses (positions and orientations) of key objects over time \cite{chi2023diffusion}. Leveraging these representations enables policies to identify relevant patterns with significantly fewer parameters. Yet, obtaining accurate 3D object poses in many scenarios remains challenging, typically requiring external tracking systems or ground-truth annotations, which limits practicality and scalability.

Recent work in learning-based 6D pose estimation and tracking, notably FoundationPose from \citet{wen2024foundationpose}, have introduced promising solutions for accurately estimating and tracking object poses from RGB-D video streams.  In theory, these methods could enable compact pose-based diffusion policy training without external tracking setups. However, a significant limitation remains: FoundationPose and similar approaches depend on manually reconstructed or hand-crafted object meshes as input. Prior attempts to integrate FoundationPose into diffusion policy frameworks \cite{hsu2024spot} have been constrained by the need for manual preprocessing of object meshes, hindering scalability in dynamic, open-set environments where object composition varies across tasks.

\begin{figure}[t]
\centering
\includegraphics[width=\columnwidth]{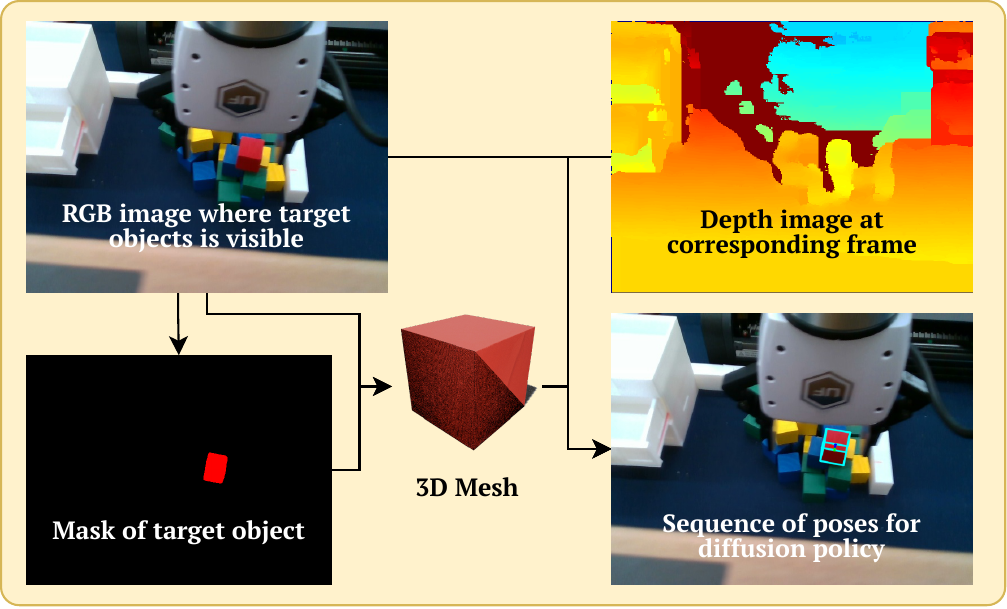}
\caption{PRISM-DP integrates state-of-the-art learning-based techniques for image segmentation, mesh generation, and unified pose estimation and tracking, enabling efficient diffusion policy learning with estimated poses from RGB-D video streams for simulation or real-world scenarios.}
\label{fig:teaser_img}
\vspace{-0.5cm}
\end{figure}

To address these challenges, we introduce \textbf{P}ose-t\textbf{R}acking via \textbf{I}mage \textbf{S}egmentation and \textbf{M}esh-generation \textbf{D}iffusion \textbf{P}olicies (PRISM-DP), a framework that leverages segmentation, mesh generation, pose estimation, and pose tracking models to enable compact diffusion policy learning directly from the spatial poses of task-relevant objects.  More specifically,  PRISM-DP applies a segmentation model to isolate task-relevant objects in input images, it uses the resulting segmentation masks to automatically create 3D triangulated meshes for all target objects using a mesh generation model, then it uses the newly created meshes to initialize a pose estimation and tracking model.  Subsequently, the diffusion policy is conditioned on both the robot state and the continuously estimated object poses for the target objects.  Crucially, because PRISM-DP uses a mesh generation model, it eliminates the need for manual mesh processing or creation, improving scalability and usability in open-set, real-world environments.

Through several experiments in both simulated and real-world settings, we demonstrate that PRISM-DP facilitates the learning of compact diffusion policies that achieve performance comparable to policies trained using ground-truth object pose information or manually generated meshes, while significantly surpassing image-based diffusion policies with similar parameter counts.  To our knowledge, our work is the first to show that diffusion policies conditioned on pose-based observations can consistently outperform those conditioned on raw image observations in a real-world setting, suggesting exciting new directions for future research.  


%% file: 2-related_works.tex
\section{Related Works}
\label{sec:related_works}

\subsection{6D Pose Estimation and Tracking}
6D pose estimation and tracking are foundational technologies in robotics, with decades of extensive study and advancement \cite{fan2022deep}. Pose estimation refers to the process of determining the 3D position and orientation (collectively known as the 6D pose) of an object from sensor data, such as images or point clouds. Traditionally, pose estimation and tracking have been treated as distinct problems. Pose estimation methods are broadly categorized into model-based approaches, which rely on known CAD or mesh models of the object \cite{labbe2022megapose, shugurov2022osop}, and model-free approaches, which instead use a few reference images of the target object \cite{cai2020reconstruct}. Although model-free methods appear more user-friendly, they often require fine-tuning for each novel object \cite{liu2022gen6d}, suffer from inefficiencies due to internal reconstruction processes \cite{he2022onepose++}, or fail under conditions of occlusion and low texture \cite{he2022fs6d}. As a result, model-based methods are generally preferred for achieving robust and reliable pose estimation.

Pose tracking, by contrast, focuses on continuously updating an object's 6D pose over time by exploiting temporal information from sequential video frames. It is the task of estimating an object's changing position and orientation across a sequence of frames, typically with an emphasis on efficiency and smoothness. Like pose estimation, tracking methods can be divided into model-free \cite{ wen2023bundlesdf} and model-based approaches \cite{ stoiber2022iterative}. 

Recently, \citet{wen2024foundationpose} introduced a unified framework called FoundationPose that integrates pose estimation and tracking into a single pipeline, offering improved performance, greater efficiency, and a more streamlined user experience.  The prototype implementation of our approach (covered below) uses the FoundationPose framework for pose estimation and tracking, allowing Diffusion Policies to train in the observation-space of key object poses.

\subsection{Policy Learning with Poses}
Pose information provides a compact and structured description of the environment, making it ideal for policy learning. However, deploying pose-based policies in the real world remains challenging due to the difficulty of acquiring accurate 6D poses.

External tracking methods such as AprilTags \cite{ angelopoulos2023high} or Motion Capture \cite{lindenheim2023yolov5} are commonly used but require specialized hardware and controlled environments. Others reconstruct scenes into point clouds or voxel maps using depth sensors \cite{li2021leveraging, wu2024deep} or LiDAR \cite{sun2023mega, sun2023benchmark}, but such methods are sensitive to environmental changes and require repeated preprocessing if changes occur. Diffusion policies initially trained with ground-truth poses in simulation have been adapted for real-world tasks via visual encoders that process input images \cite{chi2023diffusion, sun2025dynamic}, but this technique comes at the cost of three to four times increased model size.

Recent work has incorporated FoundationPose \cite{wen2024foundationpose} into learning-based robotics pipelines \cite{hsu2024spot}. However, \citet{pan2025omnimanip} use poses only for task planning, while \citet{hsu2024spot} requires manual mesh scans using consumer devices. In contrast, our approach removes the need for manual reconstruction by integrating segmentation and mesh generation for open-set 6D pose tracking. 

\subsection{3D Generation}
The success of large-scale generative models for text and images has recently extended to 3D AI-Generated Content (3D AIGC), with diffusion \cite{ho2020denoising} and Transformer-based architectures \cite{vaswani2017attention} driving progress in mesh generation. Recent methods fall into two categories: reconstruction-based approaches \cite{boss2024sf3d, xiang2024structured}, which first generate intermediate 3D representations before converting them into meshes, and autoregressive approaches \cite{chen2024meshanything, siddiqui2024meshgpt}, which directly model mesh structures through sequence learning.

Reconstruction-based methods tend to produce higher visual fidelity \cite{boss2024sf3d, xiang2024structured}, while autoregressive methods generate meshes with more concise and elegant topology \cite{chen2024meshanything, siddiqui2024meshgpt} but often at the cost of visual quality \cite{3darena}. Given the importance of geometric realism for downstream pose estimation and tracking, our approach uses a reconstruction-based mesh generation approach.

%% file: 3-technical_overview.tex
\section{Methodology}
\label{sec:technical_overview}

\begin{figure*}[t]
\centering
\includegraphics[width=\textwidth]{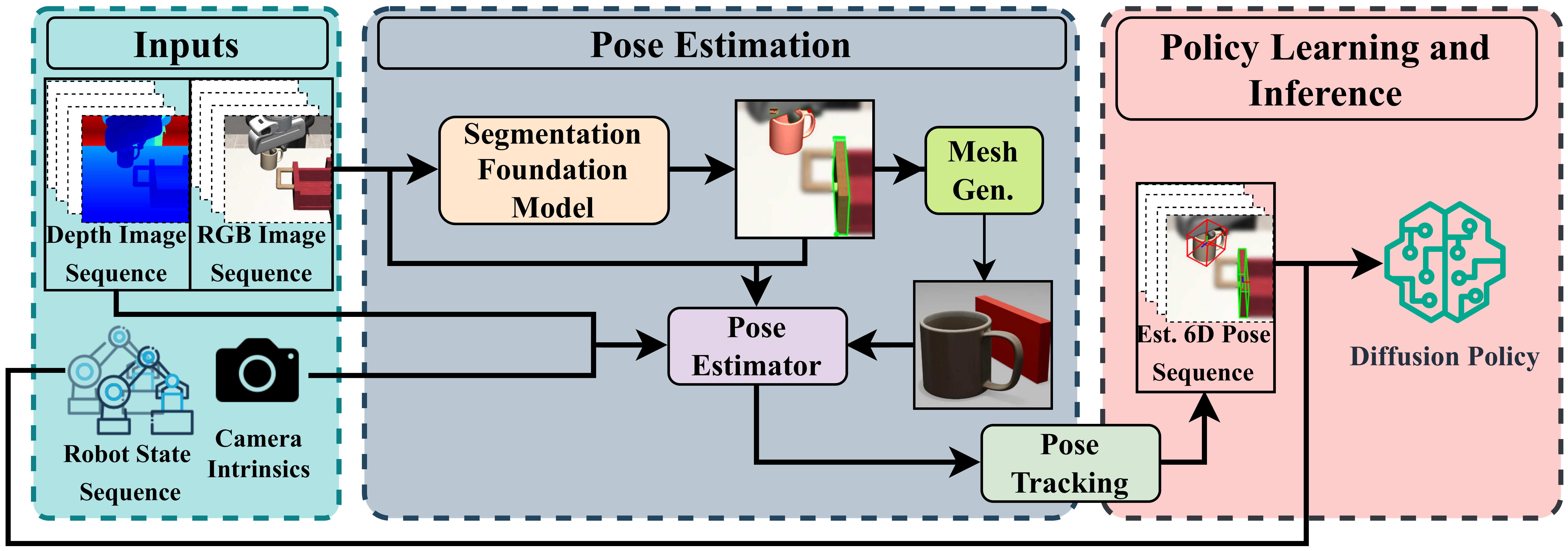}
\caption{Overview of PRISM-DP. We start from camera intrinsics $\mathbf{K}$, robot states $\mathbf{s}_t$, RGB images $\mathbf{I}_t$ and depth images $\mathbf{D}_t$ at all time steps $t$.  Segmentation, Mesh Generation, Pose Estimation, and Pose Tracking models are used to convert these inputs into a sequence of estimate poses for all target objects through a demonstration, which can then be used as observations in a Diffusion Policy.}
\label{fig:framework}
\vspace{-0.6cm}
\end{figure*}

Our goal is to learn a compact and efficient receding horizon policy \(\pi_\theta\) that, given a sequence of past observations \(\mathbf{o}_{t-H_o:t}\), predicts a future action sequence \(\mathbf{a}_{t:t+H_p}\), where \(t\) denotes the current time step, \(H_o\) is the observation horizon, and \(H_p\) is the prediction horizon. 

Commonly, an observation $\mathbf{o}_t$ at a given time $t$ in diffusion policies consists of raw RGB images, \(\mathbf{I}_t \in \mathbb{R}^{H \times W \times 3}\). However, to encourage a more compact policy in our approach, we instead reformulate  observations \(\mathbf{o}_t\) to only include low-dimensional, task-relevant information, e.g., the estimated poses of target objects.  Specifically, an observation at time \(t\) in our approach is defined as as $\mathbf{o}_t = \left[ \mathbf{s}_t, \left\{ \hat{\mathbf{T}}^t_i \right\}_{i=1}^{J} \right] $, where \(\mathbf{s}_t \in \mathbb{R}^{d_s}\) is the proprioceptive state of the robot, and \(\left\{ \hat{\mathbf{T}}^t_i \right\}_{i=1}^{J}\) is a set of $J$ separate mathematical objects, e.g., $SE(3)$ matrices or unit quaternions, specifying the estimated spatial poses of the $J$ task-relevant objects at the given time \(t\).       

The primary challenge through our work lies in inferring the $\hat{\mathbf{T}}_i^t$ poses for all $i \in \{1, ..., J\}$ and all $t \in \{1, ..., T\}$, in either simulation or real-world environments, without relying on specialized setups (e.g., markers or motion capture) or requiring an unreasonable amount of manual effort from the user.  Our approach, illustrated in Fig.~\ref{fig:framework}, addresses this problem using segmentation, mesh generation, pose estimation, and pose tracking models to conveniently automate the process of converting standard image-based inputs into a sequence of pose estimates for task-relevant objects. The approach takes four inputs: (1) a sequence of RGB images $\mathbf{I}_t \in \mathbb{R}^{H \times W \times 3}$ for all $t \in \{1, ..., T\}$; (2) a sequence of depth images $\mathbf{D}_t \in \mathbb{R}^{H \times W}$ for all $t \in \{1, ..., T\}$; (3) a sequence of robot states $\mathbf{s}_t$ for all $t \in \{1, ..., T\}$; and the camera intrinsics $\mathbf{K} \in \mathbb{R}^{3 \times 3}$.  

Using these inputs, our approach follows a five step process:

(1) For each target object $i \in \{1, \dots, J\}$, the user manually selects a frame index $F_i$ where the $i$-th target object is first clearly visible in the RGB image $\mathbf{I}_{F_i}$.  If the $i$-th object is clearly visible at the start of the image sequence, then $F_i = 1$.  Our approach currently assumes that once a target object is visible, it will remain in view throughout the rest of the demonstration.  

(2) A segmentation model produces binary masks for each target object at their respective frames of first visibility: $$\mathbf{M}_{i} = \mathrm{Segmentation}(\mathbf{I}_{F_i}), \ \forall i \in \{1, ..., J\}.$$
Each pixel in a mask $\mathbf{M}_{i} \in \mathbb{R}^{H \times W}$ is $1$ if part of the $i$-th object at time $t$ and $0$ otherwise.  

(3) Each mask $\mathbf{M}_{i}$ is passed to a mesh generation model to obtain a triangular mesh: 
$$\mathcal{M}_i = \mathrm{MeshGen}(\mathbf{I}_{F_i} * \mathbf{M}_{i}), \forall i \in \{ 1, ...,J\}.$$
The operation $\mathbf{I}_{F_i} * \mathbf{M}_{i}$ denotes elementwise multiplication, preserving RGB values in $\mathbf{I}_{F_i}$ where the mask is one and setting all other pixels to black (as seen in Fig. \ref{fig:teaser_img}). 

(4) An approximate pose is calculated at the frames of first visibility for all target objects: 
$$\hat{\mathbf{T}}_i^{F_i} = \mathrm{PoseEstimator}(\mathbf{I}_{F_i}, \mathbf{D}_{F_i}, \mathcal{M}_i, \mathbf{K}), \ \forall i \in \{1, ..., J\}.$$

(5) The pose estimates at the the frames of first visibility are used to bootstrap a pose tracking model per object: \begin{equation*}
\label{eq:pose_tracking}
\begin{split}
\hat{\mathbf{T}}_i^t &= \mathrm{PoseTracking}(\hat{\mathbf{T}}_i^{t-1}, \mathbf{I}_t, \mathbf{D}_t, \mathcal{M}_i, \mathbf{K}), \\
&\quad \forall i \in \{1, ..., J\}, \ \forall t \in \{F_i, ..., T\}.
\end{split}
\end{equation*}
If an object is not visible at the beginning of a demonstration (i.e., $F_i \neq 1$ for some $i$), then the object poses at frames preceding $F_i$ are considered null, set as $4 \times 4$ matrices of all zeros.

The output of step 5 is a set of pose estimates over all task-relevant objects and time points.  The pose estimates at time $t$ are concatenated with their corresponding robot state, $\mathbf{s}_t$, to create a sequence of observations, which are finally used to train a diffusion policy in a compact, low-dimensional space.  

Conditioned on the observation history $\mathbf{o}_{t-H_o:t}$, the diffusion policy $\pi_\theta$ outputs a predicted sequence of actions $\mathbf{a}_{t:t+H_p}$. In a general diffusion model, a forward noising process gradually corrupts clean data $\mathbf{x}_0$ with Gaussian noise over a series of timesteps. Due to the Markov property of this process, this can be reparameterized into a closed-form solution that directly samples a noisy version $\mathbf{x}_k$ at any timestep $k$:
$$
\mathbf{x}_k = \sqrt{\bar{\alpha}_k} \, \mathbf{x}_0 + \sqrt{1 - \bar{\alpha}_k} \, \boldsymbol{\varepsilon}, \quad \boldsymbol{\varepsilon} \sim \mathcal{N}(0, \mathbf{I})
$$
where $\boldsymbol{\varepsilon}$ is a standard Gaussian noise vector, and $\bar{\alpha}_k$ is a parameter from a predefined noise schedule that controls how much of the original signal $\mathbf{x}_0$ is preserved versus how much noise is added.

\begin{figure*}[t]
\centering
\begin{minipage}{\textwidth}
    
    \begin{subfigure}{\textwidth}
        \centering
        \begin{subfigure}[b]{0.195\textwidth}
            \centering
            \includegraphics[width=\linewidth]{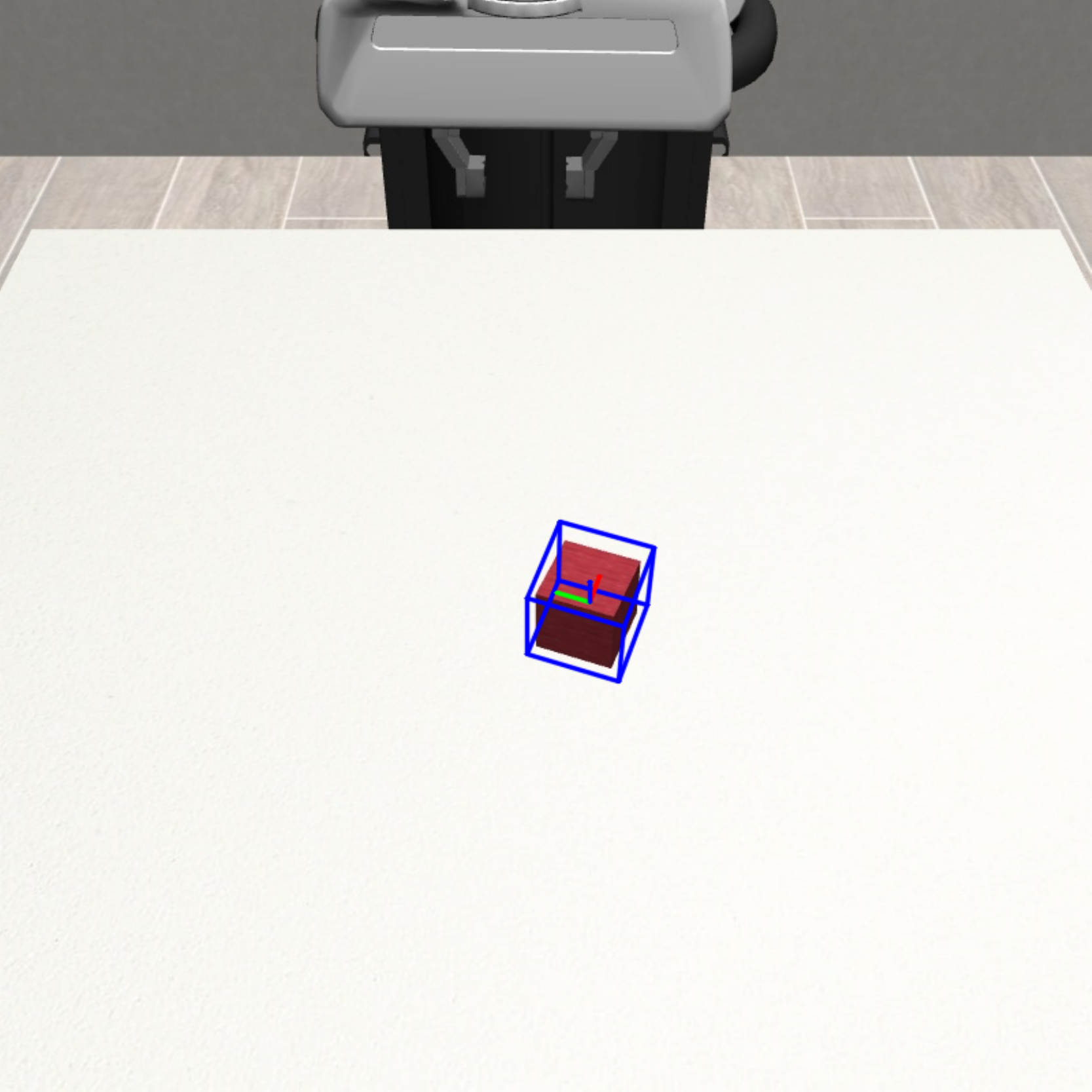}
            \caption*{Lift}
        \end{subfigure}
        \hfill
        \begin{subfigure}[b]{0.195\textwidth}
            \centering
            \includegraphics[width=\linewidth]{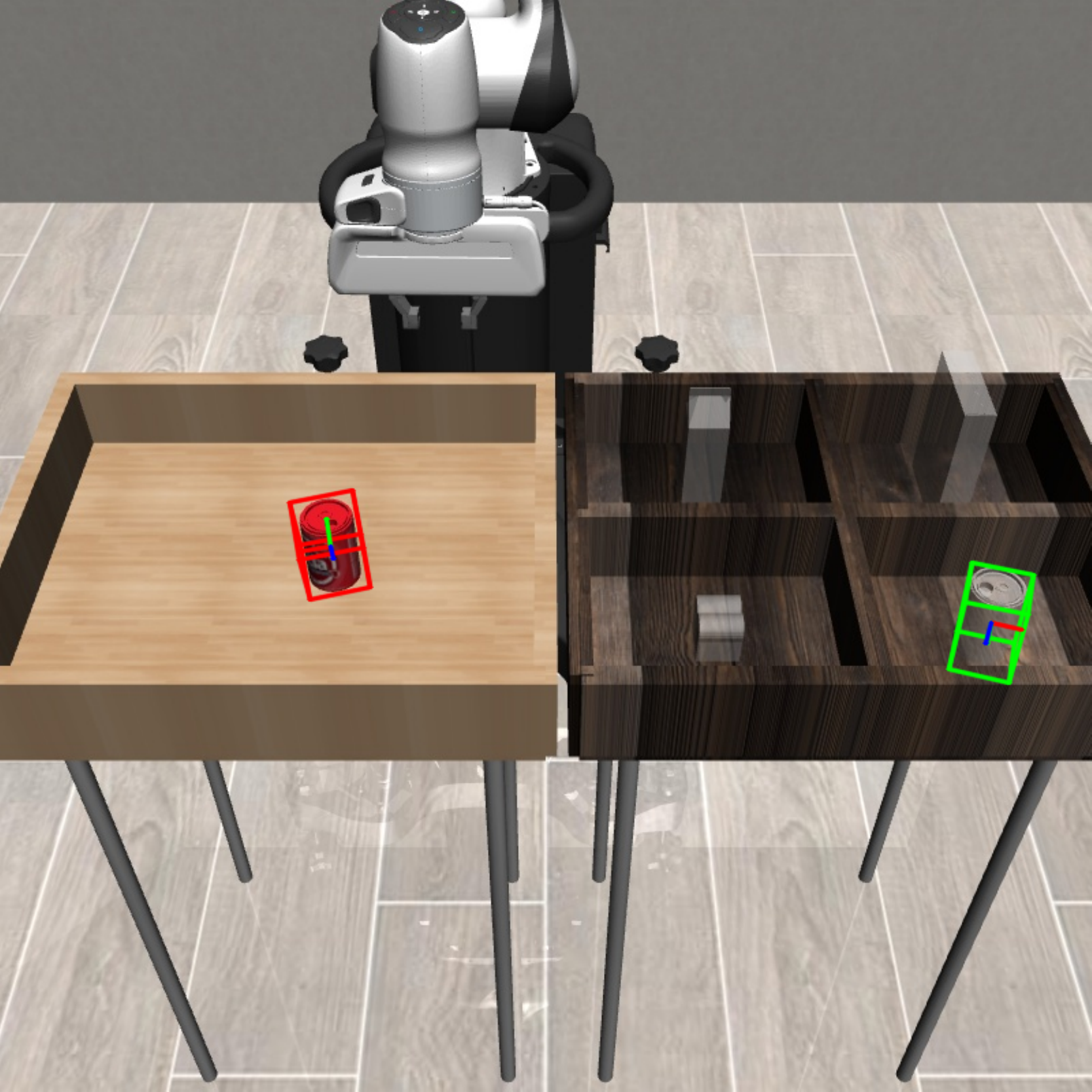}
            \caption*{Can}
        \end{subfigure}
        \hfill
        \begin{subfigure}[b]{0.195\textwidth}
            \centering
            \includegraphics[width=\linewidth]{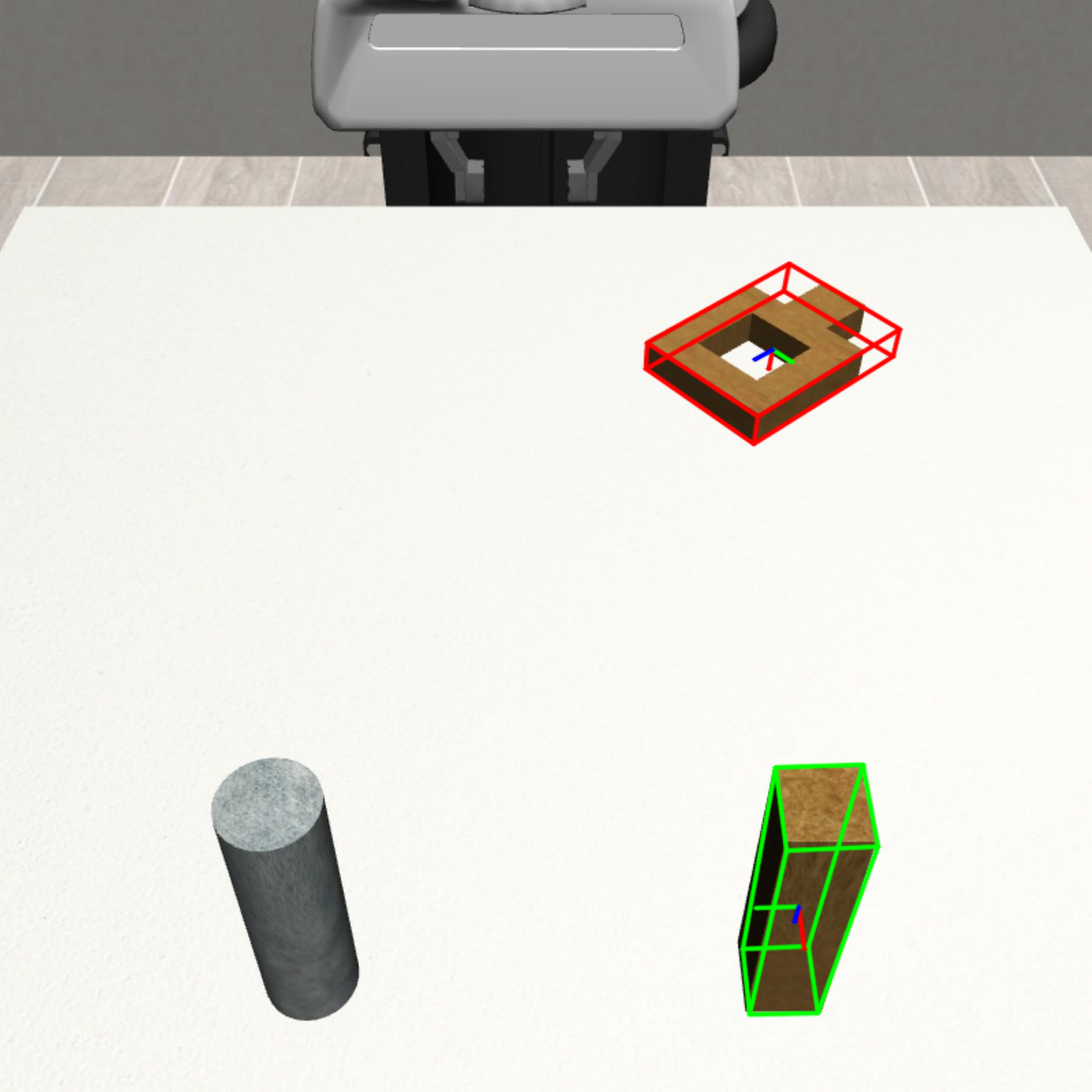}
            \caption*{Square}
        \end{subfigure}
        \hfill
        \begin{subfigure}[b]{0.195\textwidth}
            \centering
            \includegraphics[width=\linewidth]{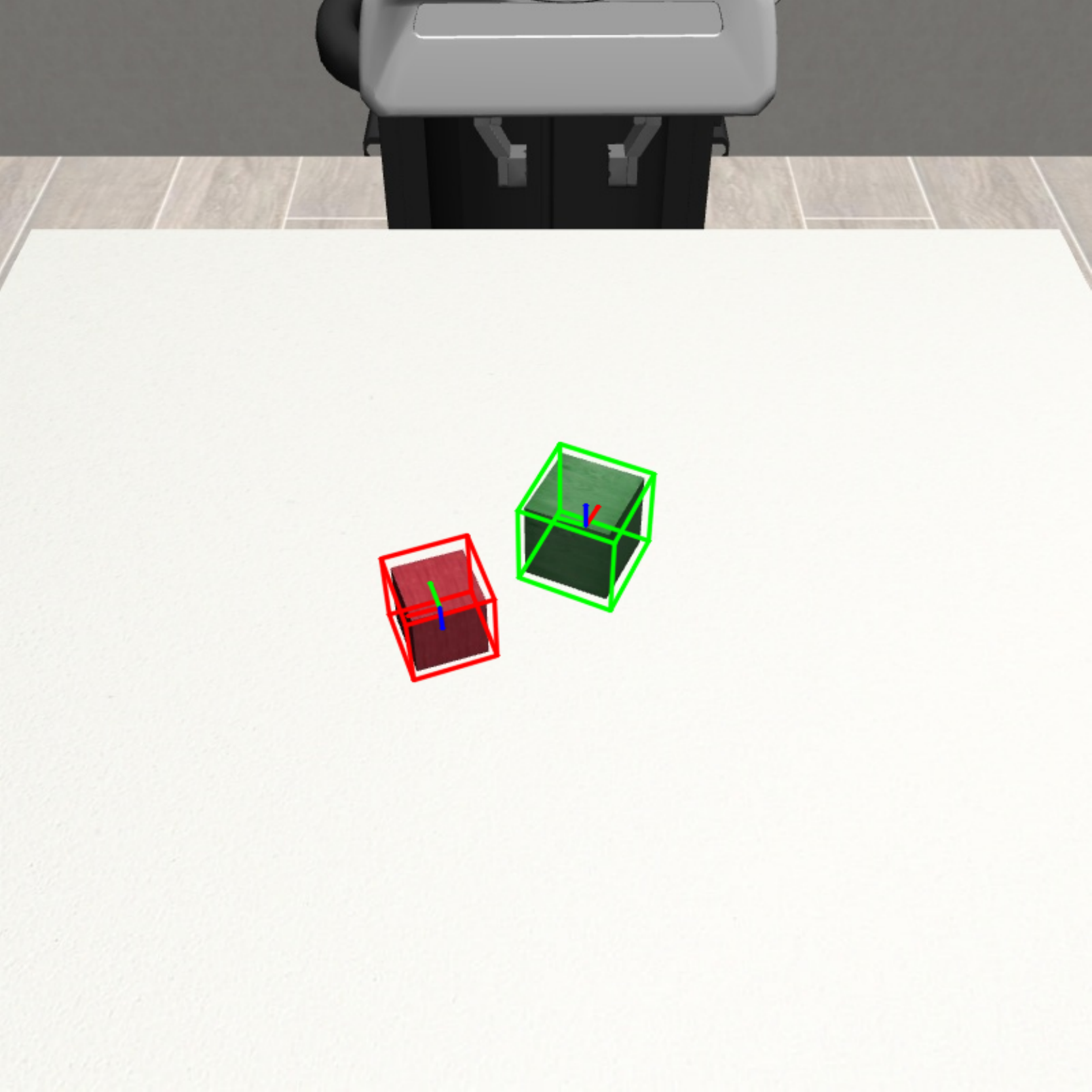}
            \caption*{Stack}
        \end{subfigure}
        \hfill
        \begin{subfigure}[b]{0.195\textwidth}
            \centering
            \includegraphics[width=\linewidth]{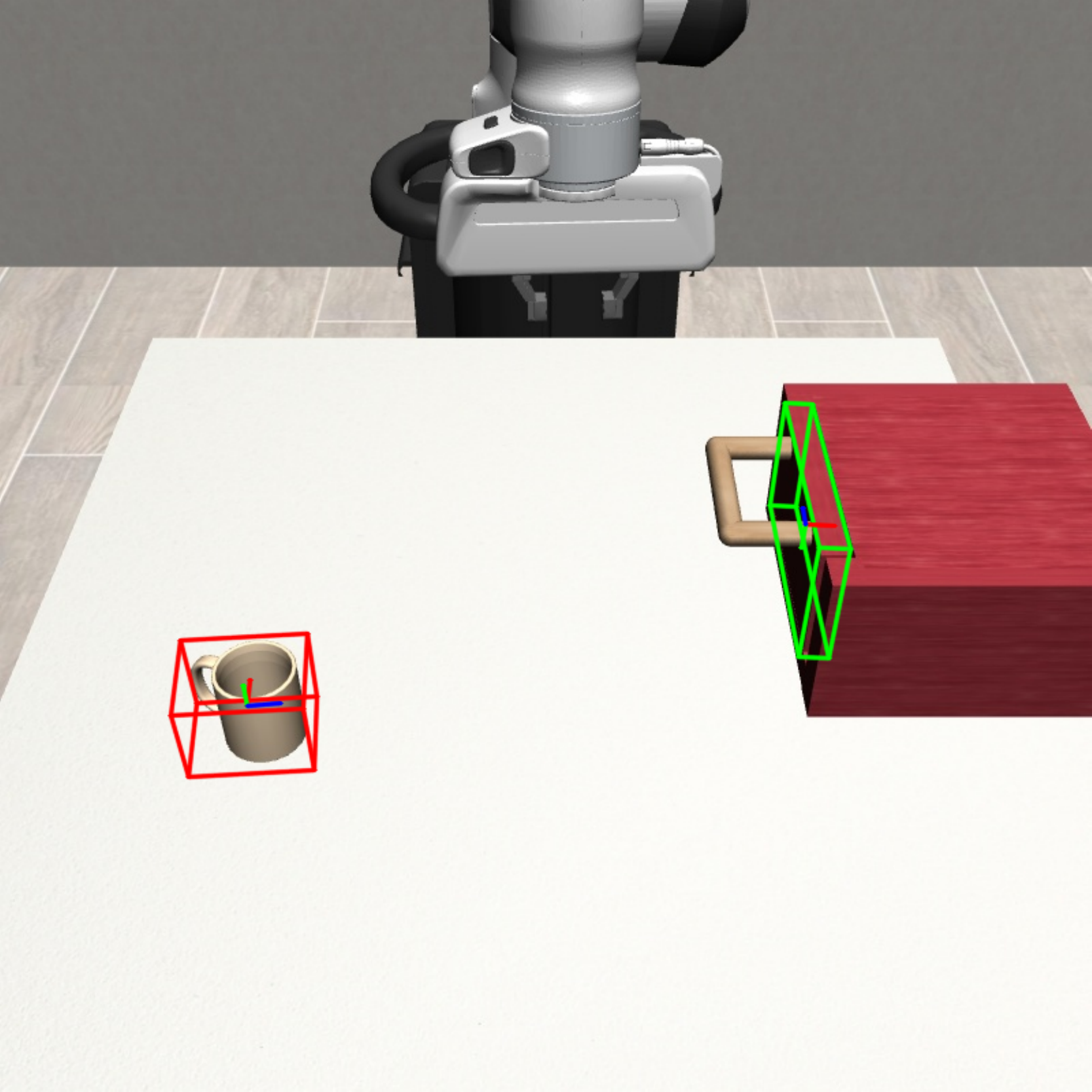}
            \caption*{Mug}
        \end{subfigure}
    \end{subfigure}
    \caption{Simulation tasks in Robosuite. Colored 3D bounding boxes indicate objects whose poses are estimated and tracked.}
    \label{fig:sim_tasks}

    \vspace{0.2cm} 

        \centering
        \captionof{table}{Success rates (SR) and training time per epoch (TE) across five Robosuite tasks. TE is recorded in seconds. Evaluated methods include Transformer-based (DP-T) and UNet-based (DP-U) diffusion policies trained with ground-truth object poses (GTP), raw RGB images (Img), and estimated 6D poses using ground-truth meshes (GTM). We also include larger image-based variants, DP-T-L Img and DP-U-L Img, that are 8 times and 11 times larger than their standard counterparts. Our proposed method with a Transformer and a UNet is denoted as PRISM-DP-T and PRISM-DP-U. }
        \label{tab:sim_eval}
        \vspace{0.2cm}
        
        \begin{tabular*}{\textwidth}{l @{\extracolsep{\fill}} *{5}{cc}}
            \toprule
            \textbf{Method} & \multicolumn{2}{c}{\textbf{Lift}} & \multicolumn{2}{c}{\textbf{Can}} & \multicolumn{2}{c}{\textbf{Stack}} & \multicolumn{2}{c}{\textbf{Square}} & \multicolumn{2}{c}{\textbf{Mug}} \\
            & \textbf{SR} & \textbf{TE} & \textbf{SR} & \textbf{TE} & \textbf{SR} & \textbf{TE} & \textbf{SR} & \textbf{TE} & \textbf{SR} & \textbf{TE} \\
            \midrule
            DP-T GTP  & $0.98$ & $0.89$ & $0.98$ & $1.87$ & $0.91$ & $2.14$ & $0.82$ & $2.88$ & $0.62$ & $9.53$ \\
            DP-U GTP  & $0.96$ & $0.87$ & $0.96$ & $2.00$ & $0.81$ & $1.95$ & $0.77$ & $2.53$ & $0.72$ & $8.23$ \\
            \midrule
            \midrule
            DP-T Img     & $0.48$ & $5.67$ & $0.25$ & $12.79$ & $0.28$ & $11.99$ & $0.08$ & $16.26$ & $0.27$ & $52.42$ \\
            DP-U Img     & $0.48$ & $5.04$ & $0.30$ & $11.27$ & $0.18$ & $10.67$ & $0.26$ & $14.40$ & $0.26$ & 46.70 \\
            DP-T-L Img   & $0.94$ & $9.32$ & $0.90$ & $21.50$ & $0.75$ & $20.30$ & $0.59$ & $27.64$ & $0.58$ & $90.79$ \\
            DP-U-L Img   & $0.94$ & $6.69$ & $0.84$ & $15.17$ & $0.74$ & $14.32$ & $0.69$ & $19.47$ & $0.54$ & 63.40 \\
            DP-T GTM    & $0.98$ & $\mathbf{0.85}$ & $\mathbf{0.98}$ & $\mathbf{1.86}$ & $\mathbf{0.86}$ & $2.18$ & 0.76 & $2.87$ & 0.64 & $9.63$ \\
            DP-U GTM    & $0.94$ & $0.87$ & $0.97$ & $1.95$ & $0.79$ & $1.90$ & $0.72$ & $\mathbf{2.65}$ & $\mathbf{0.68}$ & $8.35$ \\
            \midrule
            PRISM-DP-T      & $\mathbf{0.99}$ & $0.87$ & $\mathbf{0.98}$ & $1.87$ & $\mathbf{0.86}$ & $2.16$ & $\mathbf{0.78}$ & $2.90$ & 0.61 & $9.55$ \\
            PRISM-DP-U      & $0.95$ & $0.87$ & $0.96$ & $1.99$ & $0.80$ & $\mathbf{1.88}$ & $0.72$ & $2.50$ & $0.66$ & $\mathbf{8.33}$ \\
            \bottomrule
        \end{tabular*}
\end{minipage}
\vspace{-10pt}
\end{figure*}

During inference, the goal is to reverse this process. The model learns to gradually denoise a sample, starting from pure noise, through an iterative procedure. Specifically, at each timestep $k$, a learned denoising network $\epsilon_\theta(\mathbf{x}_k, k, \mathbf{y})$ predicts the noise component $\boldsymbol{\varepsilon}$ present in the current noisy sample $\mathbf{x}_k$, where $\mathbf{y}$ represents conditioning information. Given this predicted noise, the model constructs the mean $\mu_\theta(\mathbf{x}_k, k, \mathbf{y})$ of the reverse distribution used to sample the less-noisy version $\mathbf{x}_{k-1}$:
$$
\mu_\theta(\mathbf{x}_k, k, \mathbf{y}) = \frac{1}{\sqrt{\alpha_k}} \left( \mathbf{x}_k - \frac{1 - \alpha_k}{\sqrt{1 - \bar{\alpha}_k}} \cdot \epsilon_\theta(\mathbf{x}_k, k, \mathbf{y}) \right)
$$
where $\alpha_k$ is another parameter derived from the noise schedule. The network $\epsilon_\theta$ is trained to predict the ground-truth noise $\boldsymbol{\varepsilon}$ by minimizing the mean squared error loss, where the expectation is taken over the data distribution, noise, and timestep:
$$
\mathcal{L}(\theta) = \mathbb{E} \left\| \boldsymbol{\varepsilon} - \epsilon_\theta(\mathbf{x}_k, k, \mathbf{y}) \right\|^2
$$
In PRISM-DP, we instantiate this diffusion model as our policy $\pi_\theta$ on action sequences. Therefore, the training objective therefore becomes:
$$
\mathcal{L}_{\text{PRISM}}(\theta) = \mathbb{E} \left\| \boldsymbol{\varepsilon} - \epsilon_\theta(\mathbf{a}_{t:t+H_p, k}, k, \mathbf{o}_{t-H_o:t}) \right\|^2,
$$
where we replace the generic data $\mathbf{x}_k$ with the noisy action sequence $\mathbf{a}_{t:t+H_p, k}$ and the conditioning $\mathbf{y}$ with our observation window $\mathbf{o}_{t-H_o:t}$.

%% file: 5-evaluation.tex
\section{Evaluation}
\label{sec:evaluation}

\subsection{System Implementation}

In this subsection, we provide implementation details for our system implementation that instantiates the approach outlined in \S\ref{sec:technical_overview}.  For the $\mathrm{Segmentation}$ subroutine, the base model in our system is Segment Anything 2 (SAM2) \cite{ravi2024sam}.  Task-relevant objects are specified to SAM2 by clicking on the target objects within the sequence of RGB images on their first visible frames, $\mathbf{I}_{F_i}$.  Such inputs are called ``point prompts''.  

For the $\mathrm{MeshGen}$ subroutine, we use the Meshy model \cite{3darena, meshy_ai} due to its ability to produce high-fidelity 3D outputs. Meshy generates a mesh automatically from a masked RGB image, where the region of interest is shown in color and all other regions are blacked out. 

Lastly, for the $\mathrm{PoseEstimator}$ and $\mathrm{PoseTracking}$ subroutines, we use FoundationPose \cite{wen2024foundationpose}, which is able to provide unified pose estimation and tracking.  

During task execution, our system can run at 10 Hz control frequency constantly, introducing no practical latency bottleneck. Initially, the poses of all task-relevant objects are set to null values (all zeros). When a target object first appears in the field of view, the user provides point prompts to SAM2. FoundationPose then uses the resulting segmentation mask and corresponding mesh object (cached during training) to initialize the object's pose via pose estimation. To demonstrate the generalizability of our proposed system, all experiments are initialized from random states instead of in-distribution states.

\subsection{Simulation Experiments}

\subsubsection{Experimental Settings}
To evaluate the effectiveness of PRISM-DP, we conduct extensive experiments in simulation across five tasks from Robosuite \cite{zhu2020robosuite}, using a 7-DOF simulated Franka Panda arm. The control is executed in end-effector space, with rotations represented as quaternions. As illustrated in Fig. \ref{fig:sim_tasks}, the tasks are lifting a red cube (Lift), sorting a can into its designated compartment (Can), inserting a square nut onto a square peg (Square), stacking a red cube onto a green cube (Stack), and placing a mug into a drawer for cleanup (Mug).

For training, the datasets for Lift and Can are from Robomimic \cite{robomimic2021}, while datasets for Stack, Square, and Mug are generated using MimicGen \cite{mandlekar2023mimicgen}. All tasks are trained with 200 demonstration rollouts, except Mug, which uses 300 rollouts to account for its increased complexity.

The simulation evaluation includes a comprehensive set of baselines and architectural variants, as summarized in Table~\ref{tab:sim_eval}.  Evaluated methods include Transformer-based (DP-T) and UNet-based (DP-U) diffusion policies trained with ground-truth object poses (GTP), raw RGB images (Img), and estimated 6D poses using ground-truth meshes (GTM). We also include larger image-based variants, DP-T-L Img and DP-U-L Img, that are 8 times and 11 times larger than their standard counterparts (detailed further in the following paragraph). Our proposed method with a Transformer and a UNet is denoted as PRISM-DP-T and PRISM-DP-U. 

For policies with image input, the images have resolution of $200\times200$. Each UNet-based model uses 24 million parameters in the denoising network, except DP-U-L Img, which uses 261 million. Similarly, all Transformer-based policies use 35 million parameters, with the exception of DP-T-L Img, which scales up to 277 million to better handle high-dimensional image conditions.

All models are implemented in PyTorch \cite{NEURIPS2019_9015}. Training and evaluation of policies are performed on a workstation equipped with an AMD PRO 5975WX CPU, dual NVIDIA RTX 4090 GPUs, and 128GB of RAM. Each policy is trained for 200 epochs with batch size of 128, and evaluated over 250 rollouts to calculate the success rate by dividing the total number of evaluation rollouts with number of successful evaluation rollouts.

\subsubsection{Results} 
As shown in Table~\ref{tab:sim_eval}, PRISM-DP consistently outperforms image-conditioned diffusion policies with the same network capacity while having better training efficiency, regardless of whether a Transformer or UNet architecture is used. It achieves comparable performance to diffusion policies trained with poses obtained from ground-truth meshes, indicating that learning-based mesh generation is sufficiently accurate to support downstream pose tracking and policy learning for these tasks. While scaling up the parameter count by 8 times and 11 times in the Transformer-based and UNet-based image models (DP-T-L Img and DP-U-L Img) closes the performance gap for simpler tasks such as Lift and Can, this improvement comes at a significant computational cost and fails to generalize to more challenging tasks. For instance, in tasks like Stack, Square, and Mug, PRISM-DP still outperforms these large image-based baselines, underscoring the value of low-dimensional structured observations.

Notably, PRISM-DP also matches the performance of ground-truth pose-conditioned policies on simpler tasks with only one dynamic object (Lift and Can), but shows a slight performance drop on tasks with multiple dynamic objects (Stack, Square, Mug). This gap is expected, as these baselines have direct access to perfectly accurate object poses, whereas PRISM-DP relies on learned perception modules for object segmentation, mesh generation, and pose estimation.

\begin{figure}[t] 
    \centering

    \begin{minipage}[t]{0.49\columnwidth}
        \centering
        \includegraphics[width=\linewidth]{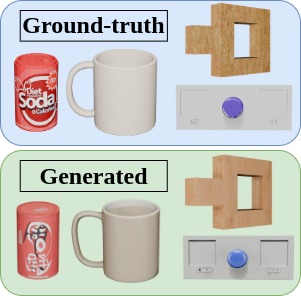}
    \end{minipage}
    \hfill 
    \begin{minipage}[t]{0.49\columnwidth}
        \centering
        \includegraphics[width=\linewidth]{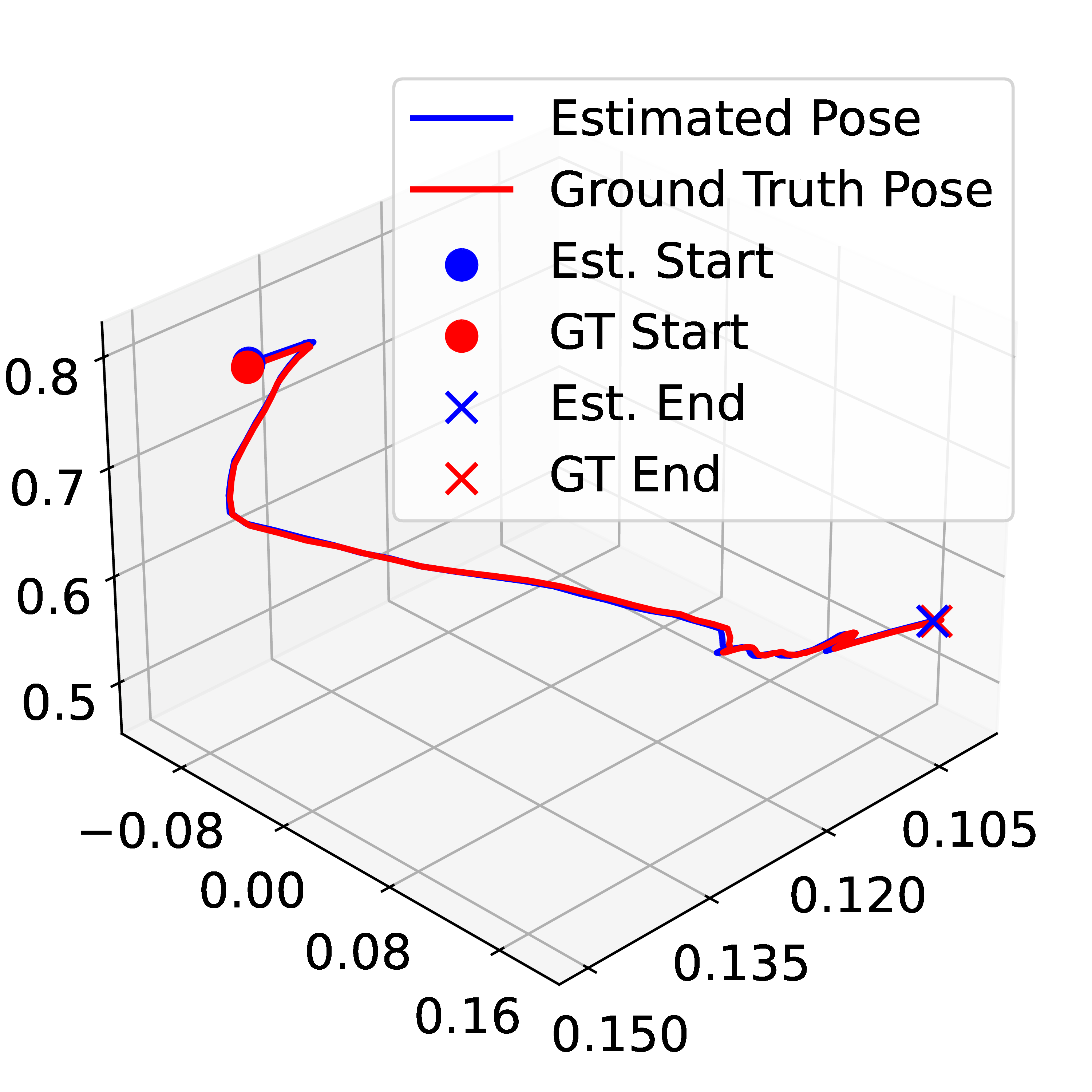}
    \end{minipage}

    \vspace{6pt} 

    \begin{minipage}{\columnwidth}
        \begin{tabular*}{\columnwidth}{@{\extracolsep{\fill}}cccc} 
            \toprule
            \textbf{PSNR} & \textbf{FID} & \textbf{Pos. Error} & \textbf{Ori. Error} \\
            \midrule
            $26.4593$ & $0.1347$ & $0.0006$ & $0.7858$ \\
            \bottomrule
        \end{tabular*}
    \end{minipage}

    \caption{Top left: Examples of ground-truth and generated meshes, shown in the upper and lower rows, respectively. Top right: Visualization of estimated and ground-truth poses of one object during a rollout. Bottom: Average PSNR, FID, and the average positional and orientation errors in radians across all objects in simulation training datasets.}
    \label{fig:pose_est_analysis}
    \vspace{-12pt}
\end{figure}

To further understand the efficacy of PRISM-DP, we evaluate the quality of mesh generation as well as the accuracy of pose estimation and tracking. For mesh generation, we compare meshes produced by the generative model against ground-truth meshes from Robosuite. For pose estimation, we compare the predicted poses, estimated using the generated meshes, to the ground-truth poses transformed into the camera frame.

To assess mesh quality, we render both ground-truth and generated meshes from 30 randomly sampled viewpoints using Blender \cite{blender}, and compute two commonly used perceptual metrics: Peak Signal-to-Noise Ratio (PSNR) \cite{huynh2008scope} and Fréchet Inception Distance (FID) \cite{heusel2017gans}. PSNR measures pixel-level reconstruction fidelity, where higher values indicate closer alignment, with 20–40 dB considered high quality \cite{rombach2022high}. FID assesses similarity in feature space, where lower scores are better, and values below 10 typically reflect high-quality generation \cite{li2024autoregressive}.

To evaluate pose estimation and tracking, we compare the estimated poses, obtained using the generated mesh as input to FoundationPose, with the corresponding ground-truth object poses in the camera frame, across all training samples. Positional accuracy is measured by the Euclidean distance between the predicted and ground-truth translations. Orientation accuracy is measured by the angular distance between predicted and ground-truth quaternions, expressed in radians. We report the mean of each error metric over all samples.

As illustrated in Fig.~\ref{fig:pose_est_analysis}, the generated meshes achieve excellent FID scores, indicating strong perceptual alignment with the ground-truth meshes. While PSNR scores are in the upper-mid range, suggesting minor pixel-level discrepancies, the low FID underscores that these differences are not semantically significant. This is particularly important given that FoundationPose is a learning-based method that relies on rendered image similarity in feature space for pose refinement. The high perceptual quality of the meshes results in extremely low position errors during pose estimation. Although the orientation error appears comparatively higher, this is largely due to consistent differences in the canonical orientation of the generated meshes versus the ground-truth CAD models. Since FoundationPose tracks pose updates relative to the given mesh geometry, these orientation offsets are typically stable and do not degrade downstream policy performance.

\subsection{Real-World Experiments}

\subsubsection{Experimental Settings}
\begin{figure}[t] 
    \centering
    \vspace{-12pt}

    \begin{minipage}{\columnwidth}
        \centering 
        \includegraphics[width=0.9\linewidth]{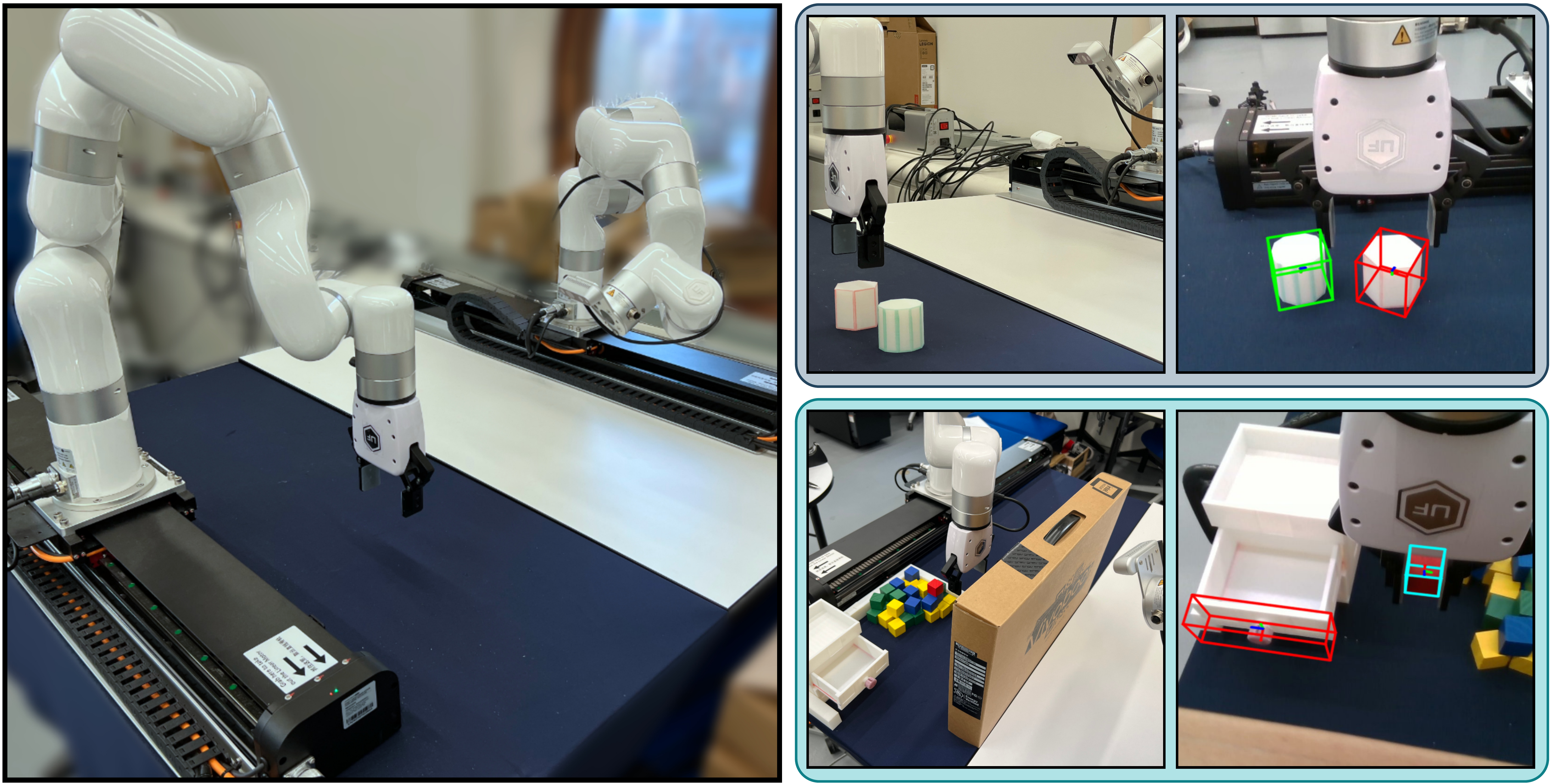}
    \end{minipage}
    
    \vspace{6pt}

    \begin{minipage}{\columnwidth}
        \begin{tabular*}{\linewidth}{l @{\extracolsep{\fill}} cccc}
        \toprule
        \textbf{Method} & \multicolumn{2}{c}{\textbf{Block Stack.}} & \multicolumn{2}{c}{\textbf{Drawer Inter.}} \\
        & \textbf{SR} & \textbf{TE} & \textbf{SR} & \textbf{TE} \\
        \midrule
        DP-T Img      & $0.00$  & $91.21$ & $0.00$  & $132.40$ \\
        DP-U Img      & $0.00$  & $88.13$ & $0.00$  & $126.48$ \\
        DP-T-L Img    & $0.87$ & $97.38$ & $0.57$ & $138.13$ \\
        DP-U-L Img    & $0.83$ & $93.01$ & $0.63$ & $131.71$ \\
        DP-T GTM      & $0.90$ & $7.60$ & $0.83$ & $9.36$ \\
        DP-U GTM      & $\mathbf{0.97}$ & $6.00$ & $0.80$ & $\mathbf{8.56}$ \\
        \midrule
        PRISM-DP-T   & $0.93$ & $7.22$ & $\mathbf{0.87}$ & $9.52$ \\
        PRISM-DP-U   & $0.90$ & $\mathbf{5.91}$ & $0.83$ & $8.57$ \\
        \bottomrule
        \end{tabular*}
    \end{minipage}

    \caption{Success rates (SR) and training time per epoch (TE) for all real-world tasks. The evaluated baselines and variants are the same as simulation experiments except for policies learned from ground-truth states since ground-truth states are no longer accessible in the real world.}
    \label{fig:realworld_exp}
    \vspace{-12pt}
\end{figure}

To further validate the proposed approach, we conduct real-world experiments using a dual-arm robotic system (Fig.~\ref{fig:realworld_exp}). The setup comprises two 7-DOF xArm7 manipulators, each mounted on a 1-DOF linear actuator. One arm is equipped with a RealSense D435i RGB–D camera for viewpoint control, while the other is fitted with a parallel gripper for manipulation. This configuration enables the policy to dynamically select viewpoints during task execution.

The system operates in a look-at end-effector space: the manipulation arm is controlled via its end-effector’s position and Euler orientation, while the viewpoint arm is controlled by its end-effector position only. The orientation of the viewpoint arm is automatically resolved through an additional IK constraint such that it always looks at the manipulation point on the other arm \cite{rakita2018autonomous, sun2024comparative}.

We evaluate our approach on two real-world tasks, visualized in Fig.~\ref{fig:realworld_exp}. The block stacking task (Block Stack.) requires placing a non-cuboid block onto another, while the drawer interaction task (Drawer Inter.) involves clearing a visual occlusion, retrieving a red cube, inserting it into a drawer, and closing the drawer. The blocks and drawer are 3D printed using meshes from \citet{lee2021beyond} and \citet{heo2023furniturebench}, which are also used as ground-truth meshes for evaluation. Each task is trained on a dataset of 200 demonstration rollouts.

The baselines and variants used in real-world evaluation are consistent with those from simulation, excluding baselines that rely on ground-truth poses, which are unavailable in real-world settings. Image-based policies take in images with resolution of $320\times240$. For variants that require ground-truth meshes, we use the same meshes employed for 3D printing. The red cube is handcrafted and textured using Blender. All policies are trained for 200 epochs with batch size of 128, and evaluated over 30 rollouts. Success rate is used as the primary evaluation metric. Training and evaluation are conducted on the same workstation used for simulation experiments.

\subsubsection{Results}.  Results for real-world experiments are reported in Fig.~\ref{fig:realworld_exp}. Consistent with simulation findings, policies trained using estimated poses from generated meshes perform on par with those using poses from ground-truth meshes, and overall outperform all other baselines in both tasks.  In contrast, small-capacity image-based diffusion policies fail completely across both tasks. This likely stems from the increased noise and complexity in real-world RGB observations, which overwhelms their limited representational capacity. Larger image-based diffusion policies achieve marginally lower performance than pose-based policies in the block stacking task, but show a more pronounced performance gap in the drawer interaction task. These results corroborate our simulation findings and reinforce the conclusion that scaling image-based policies to handle complex tasks is increasingly challenging.

%% file: 6-discussion.tex
\section{Discussion}
\label{sec:discussion}

In this work, we introduced PRISM-DP, a framework integrating segmentation, mesh generation, and pose tracking for efficient diffusion policy learning from estimated object poses. By using structured, low-dimensional pose observations instead of high-dimensional RGB images, PRISM-DP significantly improves both task performance and training efficiency. It outperforms image-based policies of a similar scale and surpasses much larger image-conditioned baselines, highlighting the value of structured representations. To our knowledge, this is the first work to show that diffusion policies conditioned on pose-based observations can consistently outperform those conditioned on raw images in real-world settings.

Despite these promising results, our approach has limitations. The generated meshes, while visually adequate, often have a suboptimal topology with excessive triangles, which may increase memory usage and latency during pose tracking. Additionally, our method is restricted to rigid-body motion and cannot model non-rigid deformations like folding cloth. Addressing these challenges will be critical for scaling this approach to more complex and dynamic environments.